\documentclass[final]{cvpr}

\usepackage{times}
\usepackage{epsfig}
\usepackage{graphicx}
\usepackage{amsmath}
\usepackage{amssymb}

\usepackage{amsfonts}
\usepackage{multirow, makecell}
\usepackage{subfigure}
\usepackage{algorithm}
\usepackage{algpseudocode}
\usepackage{csquotes}
\usepackage{comment}

\usepackage[pagebackref=true,breaklinks=true,colorlinks,bookmarks=false]{hyperref}

\def\cvprPaperID{3487} 
\def\confYear{CVPR 2021}

\begin{document}

\title{Searching by Generating: Flexible and Efficient One-Shot NAS with Architecture Generator}

\author{Sian-Yao Huang and Wei-Ta Chu\\
National Cheng Kung University, Tainan, Taiwan\\
{\tt\small \{P76084245,wtchu\}@gs.ncku.edu.tw}
}

\maketitle

\begin{abstract}
In one-shot NAS, sub-networks need to be searched from the supernet to meet different hardware constraints. However, the search cost is high and $N$ times of searches are needed for $N$ different constraints. In this work, we propose a novel search strategy called architecture generator to search sub-networks by generating them, so that the search process can be much more efficient and flexible. With the trained architecture generator, given target hardware constraints as the input, $N$ good architectures can be generated for $N$ constraints by just one forward pass without re-searching and supernet retraining. Moreover, we propose a novel single-path supernet, called unified supernet, to further improve search efficiency and reduce GPU memory consumption of the architecture generator. With the architecture generator and the unified supernet, we propose a flexible and efficient one-shot NAS framework, called Searching by Generating NAS (SGNAS). With the pre-trained supernt, the search time of SGNAS for $N$ different hardware constraints is only 5 GPU hours, which is $4N$ times faster than previous SOTA single-path methods. After training from scratch, the top1-accuracy of SGNAS on ImageNet is 77.1\%, which is comparable with the SOTAs. The code is available at: \url{https://github.com/eric8607242/SGNAS}.
\end{abstract}

\section{Introduction}
It is time-consuming and difficult to manually design neural architectures under specific hardware constraints. Neural architecture search (NAS) \cite{evolutionnas}\cite{oneshotnas}\cite{rlnas} aiming at automatically searching the best neural architecture is thus highly demanded. However, how to efficiently and flexibly determine the architectures conforming to various constraints is still very challenging \cite{onceforall}. 

The earliest NAS methods were developed based on reinforcement learning (RL) \cite{mnasnet}\cite{rlnas} or the evolution algorithm \cite{evolutionnas}. However, extremely expensive computation is needed. For example, 2,000 GPU days are needed by an RL method \cite{rlnas}, and 3,150 GPU days are needed by the evolution algorithm \cite{evolutionnas}.

\begin{figure}[t]
\begin{center}
   \includegraphics[width=1\linewidth]{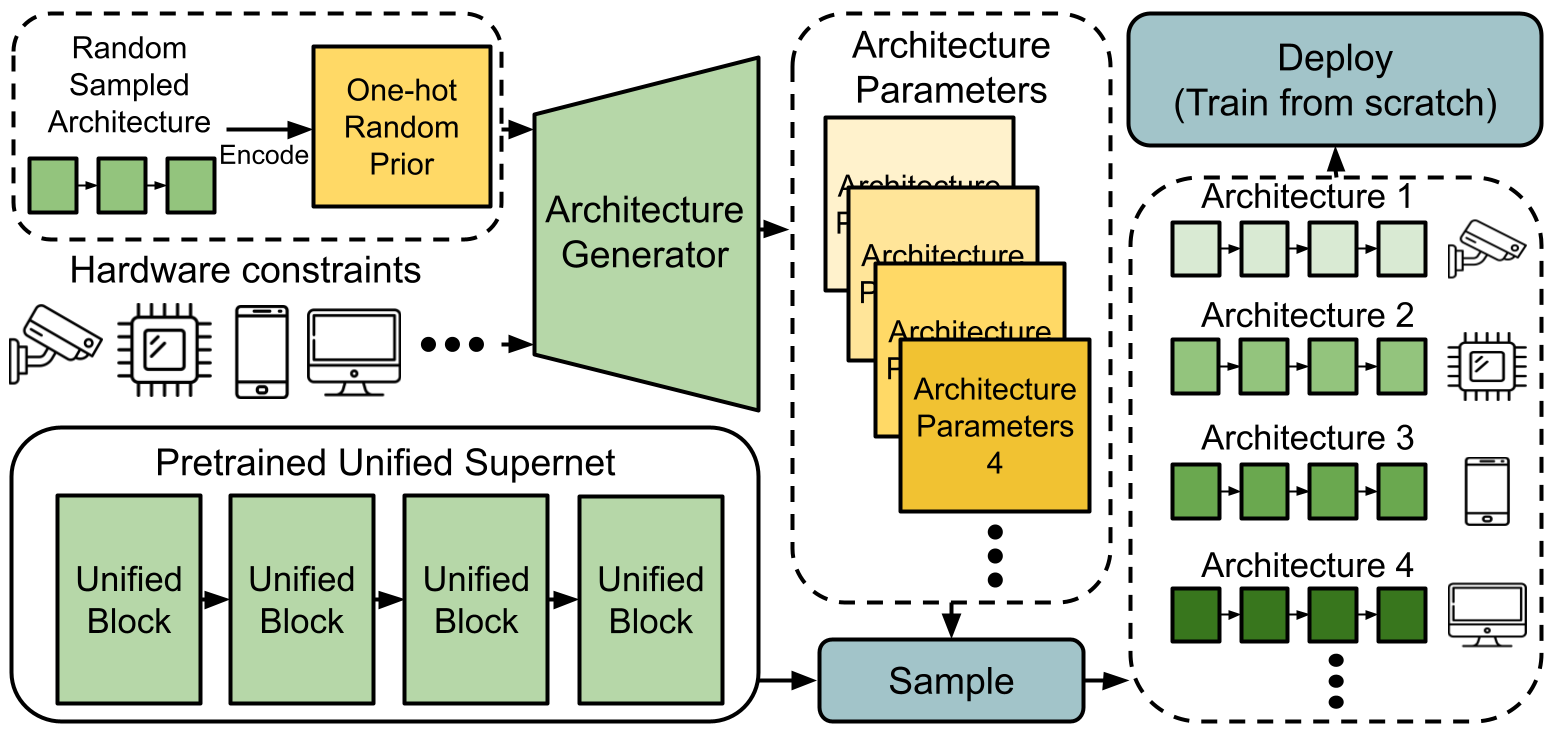}
\end{center}
   \caption{Overview of SGNAS. Given the target hardware constraint as the input, the architecture generator can generate architecture parameters instantly within the inference time of one forward pass. With the generated parameters, the specific architectures can be sampled from the unified supernet.}
\label{fig:SGNAS_overview}
\end{figure}

To improve searching efficiency, one-shot NAS methods  \cite{oneshotnas}\cite{darts}\cite{proxylessnas}\cite{fbnet} were proposed to encode the entire search space into an over-parameterized neural network, called a \emph{supernet}. Once the supernet is trained, all sub-networks in the supernet can be evaluated by inheriting the weights of the supernet without additional training. One-shot NAS methods can be divided into two categories: differentiable NAS (DNAS) and single-path NAS. 

In addition to optimizing the supernet, DNAS \cite{gdas}\cite{darts}\cite{fbnet}\cite{pcdarts:}\cite{atomnas} utilizes additional differentiable parameters, called \emph{architecture parameters}, to indicate the architecture distribution in the search space. Because DNAS couples architecture parameters optimization with supernet optimization, for $N$ different hardware constraints, the supernet and the architecture parameters should be trained jointly for $N$ times to find $N$ different best architectures. This makes DNAS methods inflexible. 

In contrast, single-path methods \cite{uniform_sampling}\cite{fairnas}\cite{scarlet}\cite{greedynas} decouple supernet training from architecture searching. For supernet training, only a single path consisting of one block in each layer is activated and is optimized in one iteration. The main idea is to simulate discrete neural architectures in the search space and save GPU memory consumption. Once the supernet is trained,  different search strategies, like the evolution algorithm \cite{greedynas}\cite{uniform_sampling}, can be used to search the architecture under different constraints without retraining the supernet. Single-path methods are thus more flexible than DNAS. However, re-executing the search strategy $N$ times for $N$ different constraints is costly and not flexible enough.

On top of one-shot NAS, we especially investigate \emph{efficiency} and \emph{flexibility}. For efficiency, we mean that, when the supernet is available, the time required to search the best architecture for a specific hardware constraint. For flexibility, we mean that, when $N$ different hardware constraints are to be met, how much total time required to search for $N$ best architectures. As a comparison instance, GreedyNAS \cite{greedynas} takes almost 24 GPU hours to search for the best neural architecture under a specific constraint. Totally $24N$ GPU hours are required for $N$ different constraints. 

In this work, we focus on improving efficiency and flexibility of the search strategy of the single-path method. The main idea is \emph{searching the best architecture by generating it}. First, we decouple supernet training from architecture searching and train the supernet as a single-path method. After obtaining the supernet, we propose to build an \emph{architecture generator} to generate the best architecture directly. Given a hardware constraint as input, the architecture generator can generate the architecture parameter within the inference time of one forward pass. This method is extremely efficient and flexible. The total search time for various hardware constraints of the architecture generator is only 5 GPU hours. Moreover, we do not need to re-execute search strategies or re-train the supernet once the architecture generator is trained. When $N$ different constraints are to be met, the search strategy only needs to be conducted once, which is more flexible than $N$ searches required in previous single-path methods \cite{fairnas}\cite{greedynas}\cite{scarlet}.

The aforementioned idea is on top of a trained supernet. However, we notice that searching on a single-path supernet still requires a lot of GPU memory and time because of the huge number of supernet parameters and complex supernet structure. Previous single-path NAS methods \cite{uniform_sampling}\cite{greedynas}\cite{scarlet} determine a block for each layer, and there may be different candidate blocks with various configurations. For example, GreedyNAS \cite{greedynas} has 13 types of candidate blocks for each layer, and thus size of the search space is $13^{L}$, where $L$ denotes the total number of layers in the supernet. Inspired by the fine-grained supernet in AtomNAS \cite{atomnas}, we propose a novel single-path supernet called \emph{unified supernet} to reduce GPU memory consumption. In the unified supernet, we only construct a block called \emph{unified block} in each layer. There are multiple sub-blocks in the unified block, and each sub-block can be implemented by different operations. By combining sub-blocks, all configurations can be described in a block. In this way, the number of parameters of the unified supernet is much fewer than previous single-path methods.

The contributions of this paper are summarized as follows. With the architecture generator and the unified supernet, we propose Searching by Generating NAS (SGNAS), which is a flexible and efficient one-shot NAS framework. We illustrate the process of SGNAS in Fig.~\ref{fig:SGNAS_overview}. Given various hardware constraints as the input, the architecture generator can generate the best architecture for different hardware constraints instantly in one forward pass. After training the best architecture from scratch, the evaluation results show that SGNAS achieves 77.1\% top-1 accuracy on the ImageNet dataset \cite{imagenet} at around 370M FLOPs, which is comparable with the state-of-the-arts of the single-path methods. Meanwhile, SGNAS outperforms SOTA single-path NAS in terms of efficiency and flexibility.

\section{Related Work}
Recently, one-shot NAS \cite{darts}\cite{fbnet}\cite{uniform_sampling} has received much attention because of reduced search cost brought by supernet. To futher reduce the search cost, a number of methods have been proposed, which can be roughly divided into two types: efficient NAS and flexible NAS. 

\subsection{Efficient NAS}
For efficiency, many methods were proposed to improve the supernet training strategy or redesign the supernet architecture. Stamoulis et al. \cite{singlelessfour} proposed a single-path supernet to encode architectures with shared convolutional kernel parameters, which reduce search cost of differential NAS. To reduce the huge cost when training on large-scale datasets, training supernet and searching on proxy datasets like CIFAR10 or part of ImageNet was proposed in \cite{fbnet}\cite{fbnetv2}\cite{tfnas}\cite{darts}\cite{cars}\cite{darts}. PC-DARTS \cite{pcdarts:} only sampled a small part of the supernet for training in each iteration to reduce computation cost. DA-NAS \cite{danas} designed a data-adaptive pruning strategy for efficient architecture search. 

\subsection{Flexible NAS}
For flexibility, in OFA \cite{onceforall} a single full network is carefully trained. Sub-networks inherit weights from the network and can be directly deployed without training from scratch. An accuracy predictor is trained after training supernet to guide the process for searching a specialized sub-network. FBNetV3 \cite{fbnetv3} trained a predictor on a proxy dataset. The accuracy predictor estimates performance of a candidate sub-network. However, it is still time-consuming to train an accuracy predictor.

In this work, we focus on improving the \emph{search strategy} in terms of both efficiency and flexibility. Note that our search strategy can be incorporated with the methods mentioned above. 

\section{Searching by Generating NAS}
\subsection{Background}

Given a supernet $A$ represented by weights $\boldsymbol{w}$, to find an architecture that achieves the best performance while meeting a specific hardware constraint, we need to find the best sub-network $a^*$ from $A$ which achieves the minimum validation loss $\mathcal{L}_{val}(a, \boldsymbol{w})$. Sampling $a$ from $A$ is a non-differentiable process. To optimize $a$ by the gradient descent algorithm, DNAS \cite{fbnet}\cite{proxylessnas}\cite{darts} relaxes the problem as finding a set of continuous architecture parameters $\boldsymbol{\alpha}$, and computes the weighted sum of outputs of candidate blocks by the Gumbel softmax function \cite{gumbel_softmax}: 
\begin{equation}
    x_{l+1} = \sum_i m^i_l\cdot b^i_l(x_l), 
    \label{eq:gradient_op1}
\end{equation}
\begin{equation}
    m^i_l = \frac{exp(\alpha^i_l+g^i_l/\tau)}{\sum^K_{k=1}exp(\alpha^k_l+g^k_l/\tau)}, 
    \label{eq:gradient_op2}
\end{equation}
where $x_l$ is the input tensor of the $l$th layer, $b_l^i$ is the $i$th block of the $l$th layer, and thus $b^i_l(x_l)$ denotes the output of the $i$th block. The term $\alpha_l^i$ is the weight of the $i$th block in the $l$th layer. The term $g^i_l$ is a random variable sampled from the Gumbel distribution $Gumbel(0, 1)$ and $\tau$ is the temperature parameter. The value $m_l^i$ is the weight for the output $b^i_l(x_l)$. 

After relaxation, DNAS can be formulated as a bi-level optimization: 
\begin{equation}
    \boldsymbol{\alpha}^* = \operatorname*{min}_{\boldsymbol{\alpha}}\mathcal{L}_{val}(\boldsymbol{w}^*, \boldsymbol{\alpha})
    \label{eq:bi-op_3}
\end{equation}
\begin{equation}
    \text{s.t. }\boldsymbol{w}^* = \operatorname*{argmin}_{\boldsymbol{w}}\mathcal{L}_{train}(\boldsymbol{w}, \boldsymbol{\alpha})
    \label{eq:bi-op_4}
\end{equation}
where $\mathcal{L}_{train}(\boldsymbol{w}, \boldsymbol{\alpha})$ is the training loss. 

Because of the bi-level optimization of $\boldsymbol{w}$ and $\boldsymbol{\alpha}$, the best architecture $\boldsymbol{\alpha}^*$ sampled from the supernet is only suitable to a specific hardware constraint. With this searching process, for $N$ different hardware constraints, the supernet and architecture parameters should be retrained for $N$ times. This makes DNAS less flexible. 

In contrast, single-path methods \cite{uniform_sampling}\cite{fairnas}\cite{scarlet}\cite{mixpath} decouple supernet training from architecture searching. For supernet training, only a single path consisting of one block in each layer is activated and is optimized in one iteration to simulate discrete neural architecture in the search space. We can formulate the process as: 
\begin{equation}
    \boldsymbol{w}^* = \operatorname*{argmin}_{\boldsymbol{w}}\mathbb{E}_{a\sim \Gamma(A)}(\mathcal{L}_{train}(\boldsymbol{w}(a)))
    \label{eq:training_supernet}
\end{equation}
where $\boldsymbol{w}(a)$ denotes the subset of $\boldsymbol{w}$ corresponding to the sampled architecture $a$, and $\Gamma(A)$ is a prior distribution of $a \in A$. The best weights $\boldsymbol{w}^*$ to be determined are the ones yielding the minimum expected training loss. After training, the supernet is treated as a performance estimator to all architectures in the search space. With the pretrained supernet weights $w^*$, we can search the best architecture $a^*$: 
\begin{equation}
    a^* = \operatorname*{argmin}_{a \in A}\mathcal{L}_{val}(\boldsymbol{w}^*(a)). 
\end{equation}
Single-path methods are more flexible than DNAS, because supernet training and architecture search are decoupled. Once the supernet is trained, for $N$ different constraints, only architecture search should be conducted for $N$ times.

In this work, we propose to decouple supernet training from architecture searching and train supernet as in single-path NAS (Eq.~(\ref{eq:training_supernet})). After supernet training, we search the best architecture by the gradient descent algorithm as in DNAS (Eq.~(\ref{eq:bi-op_3})). Instead of training architecture parameters for one specific hardware constraint, we propose a novel search strategy called \emph{architecture generator} to largely increase flexibility and efficiency.


\subsection{Architecture Generator}
\subsubsection{Essential Idea}
Given the target hardware constraint $C$, the architecture generator can generate the best architecture parameters for the hardware constraint $C$. The process of the architecture generator can be described as $\boldsymbol{\alpha} = G(C)$ such that $Cost(\boldsymbol{\alpha}) < C$. With the architecture generator $G$, the objective function of the architecture searching in Eq.~(\ref{eq:bi-op_3}) can be reformulated as :
\begin{equation}
    G^* = \operatorname*{min}_{G}\mathcal{L}_{val}(\boldsymbol{w}^*, \boldsymbol{\alpha}). 
    \label{eq:generator_ce}
\end{equation}


To make $G^*$ generate the best architecture parameters for different hardware constraints accurately, we propose the hardware constraint loss $\mathcal{L}_C$ as:
\begin{equation}
    \mathcal{L}_C(\boldsymbol{\alpha}, C) = (Cost(\boldsymbol{\alpha}) - C)^2,  
\end{equation}
where the cost yielded by the generated architecture $Cost(\boldsymbol{\alpha})$ is estimated by: 
\begin{equation}
   Cost(\boldsymbol{\alpha}) = \sum_{l}\sum_{i}m^i_l\cdot Cost(b^i_l). 
   \label{eq:hc_count}
\end{equation}
The term $Cost(b^i_l)$ is the constant cost of the $i$th block in the $l$ layer and $m^i_l$ is the weight of different blocks described in Eq.~(\ref{eq:gradient_op2}). The cost $Cost(\boldsymbol{\alpha})$ is differentiable with respect to $m^i_l$ and $\boldsymbol{\alpha}$, similarly in \cite{fbnet}\cite{tfnas}. Note that Eq.~(\ref{eq:hc_count}) is also highly correlated with latency, as 
mentioned in \cite{fbnet} and \cite{tfnas}. By combining the hardware constraint loss $\mathcal{L}_C$ and the cross entropy loss $\mathcal{L}_{val}$ defined in Eq.~(\ref{eq:generator_ce}), the overall loss of the architecture generator $\mathcal{L}_G$ is: 
\begin{equation}
    \mathcal{L}_G = \mathcal{L}_{val}(\boldsymbol{w}^*, \boldsymbol{\alpha}) + \lambda\mathcal{L}_C(\boldsymbol{\alpha}, C). 
    \label{eq:loss_g_1}
\end{equation}
where $\lambda$ is a hyper-parameter to trade-off the validation loss and hardware constraint loss.

\subsubsection{Accurate Generation with Random Prior}
\label{sec:randomprior}
In practice, we found that the architecture generator easily overfits to a specific hardware constraint. The reason is that it is too difficult to generate complex and high-dimensional architecture parameters based on a given simple integer hardware constraint $C$.

To address this issue, a prior is given as input to stabilize the architecture generator. We randomly sample a neural architecture from the search space, and encode the neural architecture into a one-hot vector to be the prior knowledge of architecture parameters. We name it as a random prior $B = {B_1, ..., B_L}$. Formally, $B_l = one\_hot(a_l)$, $l = 1, ..., L$, where $a_l$ is the $l$th layer of the neural architecture $a$ randomly sampled from $A$, and $L$ is the total number of layers in the supernet. With the random prior, the architecture generator is to learn the residual from the random prior to the best architecture parameters, making training architecture generator more stable and accurately (blue line in Fig.~\ref{fig:generator_random prior}(a)), and the process of the architecture generator can be reformulated as $\boldsymbol{\alpha} = G(C, B)$ such that $Cost(\boldsymbol{\alpha}) < C$. 

\subsubsection{The Generator Training Algorithm}
We illustrate the algorithm of architecture generator training in Algorithm~\ref{alg:training_generator}. In each iteration, given the target constraint and the random prior, the architecture generator can generate the architecture parameters $\boldsymbol{\alpha}$ (as illustrated in Fig.~\ref{fig:SGNAS_overview}). With $\boldsymbol{\alpha}$, the corresponding cost $C_{\boldsymbol{\alpha}}$ can be calculated by Eqn.~(\ref{eq:hc_count}). We can predict $\hat{y}$ based on the pretrained supernet $N$ with $\boldsymbol{\alpha}$. The total loss is given by Eqn.~(\ref{eq:loss_g_1}). No matter what constraint is given, the architecture generator generates architecture parameters to get the best prediction results. Therefore, training the generator is equivalent to searching the best architectures for various constraints in the proposed SGNAS. 

\subsubsection{The Architecture of the Generator}
\begin{figure}[t]
\begin{center}
   \includegraphics[width=1\linewidth]{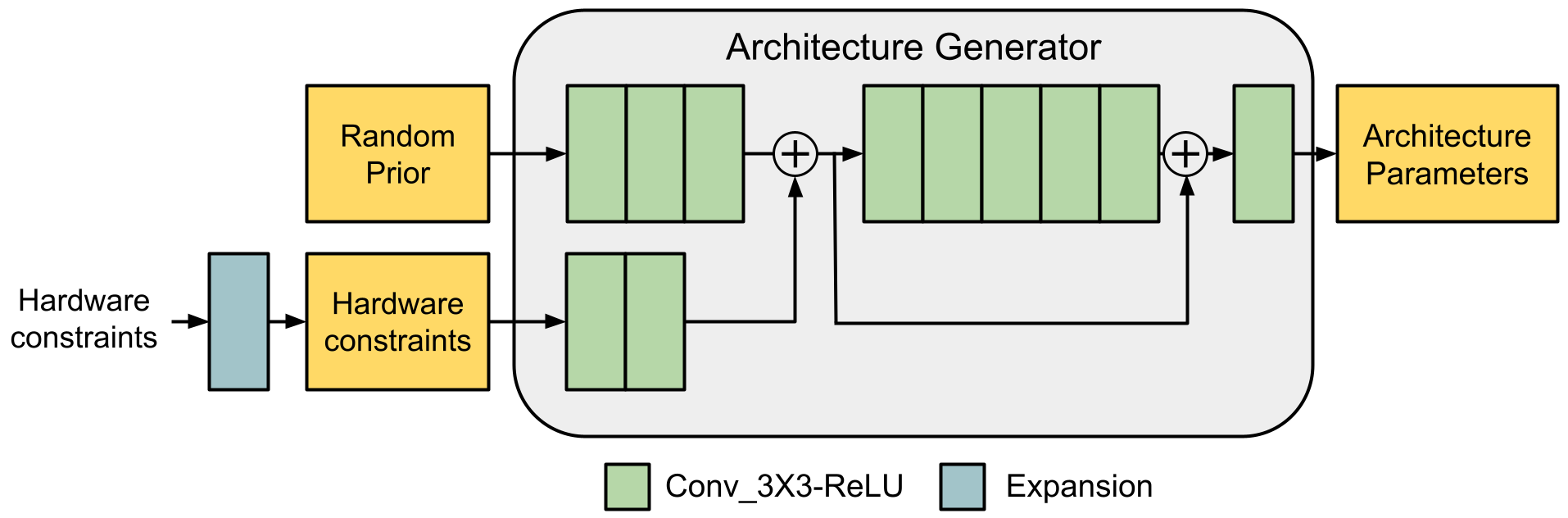}
\end{center}
   \caption{Structure of the architecture generator. Given the input target hardware constraint, the expansion layer expands the input to a tensor with shape same as the random prior.}
\label{fig:generator_structure}
\end{figure}

Fig.~\ref{fig:generator_structure} illustrates the architecture of the generator. We set the channel size of all convolutional layers as 32 and set the stride as 1, making sure the output shape same as the shape of the random prior. Please refer to supplementary materials for detailed configurations of the architecture generator and random prior representations.

\begin{algorithm}[h]
  \caption{Training Architecture Generator}
  \label{alg:training_generator}
  \begin{algorithmic}[1]
    \Require
      $B$: Random prior;
      $N$: Unified supernet;
      $G$: Generator;
      $[C_L, C_H]$: Pre-define hardware constraint interval;
      $D_{val}$: Validation dataset;
      $T$: Max iterations;
    \For{$t=1,...,T$};
      \State Get a data batch $X$ and $y$ from $D_{val}$
      \State Randomly sample $C_{target}$ from $[C_L, C_H]$
      \State $\boldsymbol{\alpha} = G(B, C_{target})$
      \State $C_{\boldsymbol{\alpha}} = Cost(\boldsymbol{\alpha})$
      \State $\hat{y} = N(X, \boldsymbol{\alpha})$ 
      \State $\mathcal{L}_{total} = \mathcal{L}_{val}(y, \hat{y})  + \lambda\mathcal{L}_C(C_{target}, C_{\boldsymbol{\alpha}})$
      \State Calculate gradients from $\mathcal{L}_{total}$
      \State Update $G$ from gradients
    \EndFor
  \end{algorithmic}
\end{algorithm}

\subsection{Unified Supernet}
Previous single-path NAS \cite{fairnas}\cite{scarlet}\cite{uniform_sampling}\cite{greedynas} adopts the MobilenetV2 inverted bottleneck \cite{mobilenetv2} as the basic building block. Given the input tensor $X$, the corresponding output $Y_{out}$ is obtained by 
\begin{equation}
    Y_{out} =  P^{c_3, c_2}(D_{K\times K}(P^{c_2, c_1}(X))), 
    \label{eq:basic_block}    
\end{equation}
where $P^{c_j,c_i}$ denotes the pointwise convolution with the input channel size $c_i$ and output channel size $c_j$, $D_{K\times K}$ denotes the depthwise convolution with $K\times K$ kernel size.
Eq.~(\ref{eq:basic_block}) represents that $X$ of $c_1$ channels is first expanded to a tensor of $c_2$ channels, which can be described as $c_2 = e \times c_1$, and then a depthwise convolution is conducted. The term $e$ denotes the expansion rate of the inverted bottleneck. After that, the tensor of $c_2$ channels is embedded into the output tensor of $c_3$ channels. Because one basic building block can only represent one configuration with one kernel size and one expansion rate, previous single-path NAS \cite{proxylessnas}\cite{scarlet}\cite{fairnas} needs to construct blocks of various configurations in each layer, which leads an exponential increase in parameter numbers and complexity of the supernet. 

In this work, we propose a novel single-path supernet called \emph{unified supernet} to improve efficiency and flexibility of the architecture generator. The only type of block, i.e., \emph{unified block}, is constructed in each layer. The unified block is built with only the maximum expansion rate $e_{max}$, i.e., $c_2 = e_{max} \times c_1$.

\begin{figure}
\begin{center}
   \includegraphics[width=1\linewidth]{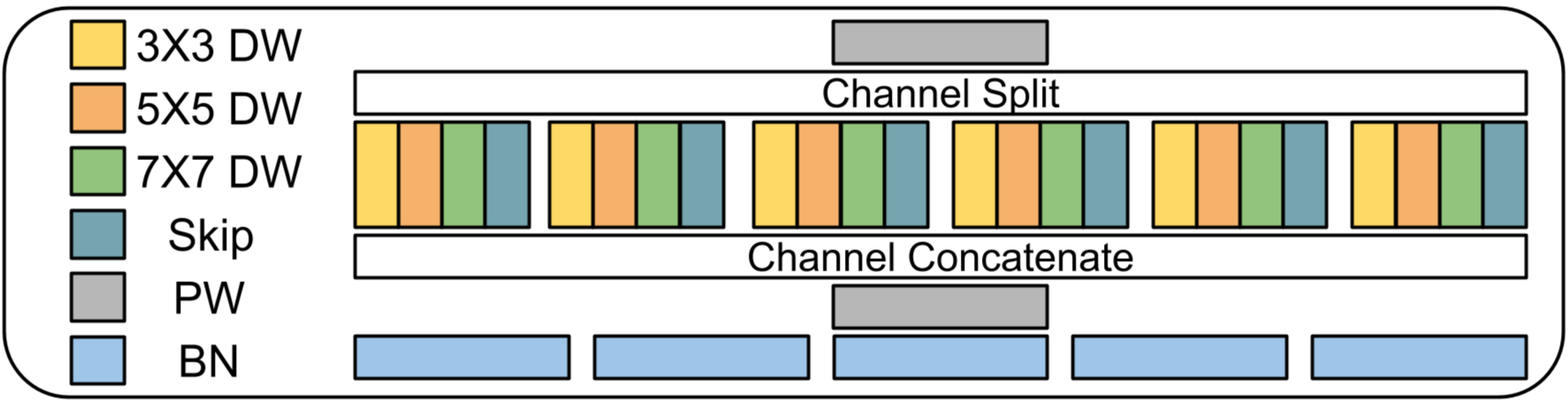}
\end{center}
   \caption{Illustration of a unified block. Given the input, the first pointwise convolution expands the input channel size to $e_{max}$ times. The channel split layer splits the tensor into $e_{max}$ parts and feeds to each sub-block, respectively. }
\label{fig:SGNAS_supernet}
\end{figure}

Fig.~\ref{fig:SGNAS_supernet} illustrates the idea of a unified block. To make the unified block represent all possible configurations, we replace the depthwise convolution $D$ by $e_{max}$ sub-blocks, and each sub-block can be implemented by different operations or skip connection. The output tensor of the first pointwise convolution $Y_1$ is equally split into $e_{max}$ parts, $Y_{1,1}, Y_{1,2}, ..., Y_{1,e_{max}}$. With the sub-blocks $d_i$ and split tensors $Y_{1,i}$, we can reformulate $D_{K \times K}$ in Eq.~(\ref{eq:basic_block}) as:
\begin{equation}
    d_1(Y_{1,1}) \circ d_2(Y_{1,2}) \circ \cdots \circ d_{e_{max}}(Y_{1,e_{max}}), 
\end{equation}
where $\circ$ denotes the channel concatenation function. 

\begin{figure}
\begin{center}
   \includegraphics[width=1\linewidth]{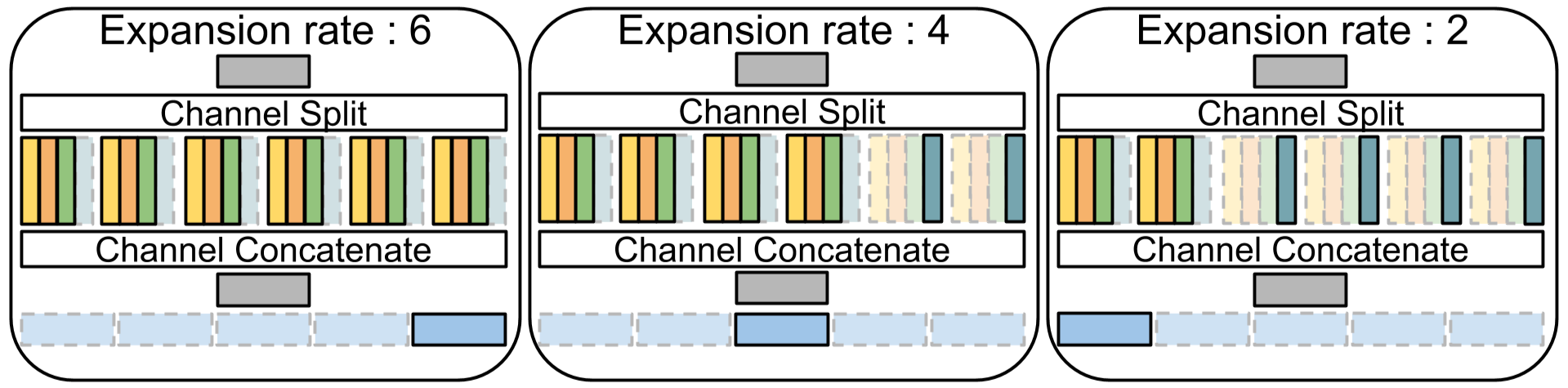}
\end{center}
   \caption{Different expansion rates can be simulated by sub-blocks with different operations. For example, the expansion rate 6 can be simulated if no skip connection is implemented, and the expansion rate 2 can be simulated if four skip connections are implemented. }
\label{fig:expansion_simulation}
\end{figure}

With the sub-blocks implemented by different operations, we can simulate blocks with various expansion rates, as shown in Fig.~\ref{fig:expansion_simulation}. The unified supernet thus can significantly reduce the parameters and GPU memory consumption. It is interesting that the MixConv in MixNet \cite{mixconv} is a special case of our search space if different sub-blocks are implemented by different kernel sizes. 


\subsubsection{Large Variability of BNs Statistics}
As in \cite{atomnas}, \cite{universlimm}, and \cite{aows}, we suffer from the problem of unstable running statistics of batch normalization (BN). In the unified supernet, because one unified block would represent different expansion rates, the BN scales change more dramatically during training. To address the problem, BN recalibration \cite{atomnas}\cite{universlimm}\cite{aows} is used to recalculate the running statistics of BNs by forwarding thousands of training data after training. On the other hand, shadow batch normalization (SBN) \cite{mixpath} or switchable batch normalization \cite{slimmable} are used to stabilize BN. In this work, we utilize SBN to address the large variability issue, as illustrated in Fig.~\ref{fig:expansion_simulation}. In our setting, there are five different expansion rates, i.e., 2, 3, 4, 5, and 6. We thus take five BNs after the second pointwise convolution block to capture the BN statistics for different expansion rates. With SBN, we can capture different statistics and make supernet training more stable. 

\subsubsection{Architecture Redundancy}
Denote two sub-blocks as $b_1$ and $b_2$. In the unified supernet, for example, the case of $b_1$ using $3 \times 3$ kernel size and $b_2$ using skip connection is distinct from the case of $b_1$ using skip connnection and $b_2$ using $3 \times 3$ kernel size. However these two cases actually correspond to the same sub-network, and thus the architecture redundancy problem arises. This redundancy makes the unified supernet more complex and hard to train. To address this issue, we force that skip connection can only be used in sub-blocks with higher index. For example, if we want to train a unified block with expansion rate 3, only the last three sub-blocks can be skip connection. We call this strategy forced sampling (FS). Please refer to supplementary materials for details of architecture redundancy and forced sampling.

\section{Experiments}
\subsection{Experimental Settings}
We adopt the macro structure of supernet (e.g., channel size of each layer and layer number) same as \cite{fairnas} and \cite{proxylessnas}, but utilize the proposed unified blocks to reduce GPU memory consumption and the number of parameters. Each sub-block in the unified blocks can be realized based on convolutional kernel sizes 3, 5, or 7, or the skip connection. We set the minimum and maximum expansion rates as 2 and 6, respectively. The size of our search space is $80^{19}$. Please refer to supplementary materials for more details. 

For experiments on the ImageNet dataset \cite{imagenet}, we train the unified supernet for 50 epochs using batch size 256 and adopt the stochastic gradient descent optimizer. 
The learning rate is decayed with the cosine annealing strategy \cite{cosine_ann} from the initial value 0.045. After supernet training, the architecture generator is trained for 50 epochs by the Adam optimizer with the learning rate 0.001.
After searching/generating the best architecture under hardware constraints, we adopt the RMSProp optimizer with 0.9 momentum \cite{mnasnet} to train the searched architecture from scratch. Learning rate is increased from 0 to 0.16 in the first 5 epochs with batch size 256, and then decays 0.03 every 3 epochs. 

\subsection{Experiments on ImageNet}
\subsubsection{Comparison with Baselines}
Li and Talwalkar \cite{randomsearch} presented that a random search approach usually achieves satisfactory performance. To make comparison, we randomly select 1,000 candidate architectures with FLOPs under 320 millions (320M) from the unified supernet and pick the architecture yielding the highest top-1 accuracy, as mentioned in \cite{randomsearch}. Besides, we also search the network with FLOPs under 320M by the evolution algorithm \cite{uniform_sampling} as another baseline. 

Table~\ref{table:baseline} shows the comparison results. As can be seen, with around 320M FLOPs, the proposed SGNAS achieves the highest top-1 accuracy. Both baselines take around 34 GPU hours to complete the search. For $N$ different hardware constraints, the search strategy should be re-executed for $N$ times, and the search time of each of two baselines is $34N$ GPU hours totally. In contrast, SGNAS only takes 5 GPU hours totally for $N$ different hardware constraints, which is much more efficient and flexible than the baselines.

\begin{table}[]
\caption{Performance comparison with baselines.}
\centering
\vspace{1mm}
\begin{tabular}{c|l|l|c|c|c}
\multicolumn{3}{c|}{Search Strategy} & \begin{tabular}[c]{@{}c@{}}Search Time\\ (GPU hrs)\end{tabular} & \begin{tabular}[c]{@{}c@{}}FLOPs\\ (M)\end{tabular} & \begin{tabular}[c]{@{}c@{}}Top-1\\ (\%)\end{tabular} \\ \hline
\multicolumn{3}{c|}{Random search} & $34N$ & 322 & 74.63 \\
\multicolumn{3}{c|}{Evolution algorithm} & $34N$ & 318 & 74.67 \\ \hline
\multicolumn{3}{c|}{\textbf{SGNAS}} & \textbf{5} & 324 & \textbf{74.87}
\end{tabular}
\label{table:baseline}
\end{table}

\subsubsection{Comparison with SOTAs}

\begin{table}[]
\centering
\caption{Comparison with the SOTAs for different hardware constraints. $^\dagger$: training with AutoAugment \cite{autoaugment}. $^\ddagger$: searching on a proxy dataset. The unit of search time and train time is GPU hours. }
\vspace{1mm}
\begin{tabular}{l|c|c|c|c}
Method & \begin{tabular}[c]{@{}c@{}}FLOPs\\ (M)\end{tabular} & \begin{tabular}[c]{@{}c@{}}Top-1\\ (\%)\end{tabular} &
\begin{tabular}[c]{@{}c@{}}Train\\ time\end{tabular} &
\begin{tabular}[c]{@{}c@{}}Search \\ time\end{tabular} \\ \hline
MobileNetV2 \cite{mobilenetv2} & 300 & 72.0 & -- & -- \\\hline
EfficientNet B0 \cite{efficientnet} & 390 & 76.3 & -- & -- \\
MixNet-M \cite{mixconv} & 360 & 77.0 & -- & -- \\
MixPath-A \cite{mixpath} & 349 & 76.9 & 240 & -- \\
AtomNAS-C \cite{atomnas} & 363 & 77.6 & 0 & $816N$\\
PC-DARTS \cite{pcdarts:} & 597 & 75.8 & 0  & $91N$ \\
ScarletNAS-A \cite{scarlet} & 365 & 76.9 & 240 & $48N$ \\
GreedyNAS-A \cite{greedynas} & 366 & 77.1 & 168 & $\sim 24N$ \\
\textbf{SGNAS-A (Ours)} & 373 & 77.1 & 280 & \textbf{5} \\ \hline
FBNetV2-L1 \cite{fbnetv2} & 325 & 77.2 & 0 & $600N^\ddagger$\\
Proxyless-R \cite{proxylessnas} & 320 & 74.6 & 0 & $200N$\\ 
FairNAS-C \cite{fairnas} & 325 & 76.7$^\dagger$ & 240 & $48N$\\ 
ScarletNAS-B \cite{scarlet} & 329 & 76.3 & 240 & $48N$ \\
SPOS \cite{uniform_sampling} & 326 & 74.5 & 288 & $\sim 24N$ \\
GreedyNAS-B \cite{greedynas} & 324 & 76.8 & 168  & $\sim 24N$ \\
\textbf{SGNAS-B (Ours)} & 326 & 76.8 & 280 & \textbf{5} \\ \hline
MobileNetV3-L \cite{mobilenetv3} & 219 & 75.2 & -- & -- \\
ScarletNAS-C \cite{scarlet} & 280 & 75.6 & 240 & $48N$ \\
GreedyNAS-C \cite{greedynas} & 284 & 76.2 & 168 & $\sim 24N$ \\
\textbf{SGNAS-C (Ours)} & 281 & 76.2 & 280  & \textbf{5}
\end{tabular}
\label{table:sota}
\end{table}

This section is dedicated to compare with various SOTA one-shot NAS methods that utilize the augmented techniques (e.g., Swish activation function \cite{swish} and Squeeze-and-Excitation \cite{se}). We directly modify the searched architecture by replacing all ReLU activation with H-Swish \cite{mobilenetv3} activation and equip it with the squeeze-and-excitation module as in AtomNAS \cite{atomnas}. 

For comparison, similar to the settings in ScarletNAS \cite{scarlet} and GreedyNAS \cite{greedynas}, we search architectures under 275M, 320M, and 365M FLOPs, and denote the searched architecture as SGNAS-C, SGNAS-B, and SGNAS-A, respectively. The comparison results are shown in Table~\ref{table:sota}. The column \enquote{Train time} denotes that the time needed to train the supernet, and the column \enquote{Search time} denotes that the time needed to search the best architecture based on the pre-trained supernet. Because DNAS couples architecture searching with supernet optimization, we list the time needed for the entire pipeline in the \enquote{Search time} column. As can be seen, our SGNAS is competitive with SOTAs in terms of top-1 accuracy under different FLOPs. For example, SGNAS-A achieves 77.1\% top-1 accuracy, which outperforms ScarletNAS \cite{scarlet} by 0.2\%, outperforms MixNet-M \cite{mixconv} by 0.1\%, outperforms MixPath-A \cite{mixpath} by 0.2\%, and is comparable with GreedyNAS-A \cite{greedynas}. 

More importantly, SGNAS achieves much higher search efficiency. With the architecture generator and the unified supernet, even for $N$ different architectures under $N$ different hardware constraints, totally only 5 GPU hours are needed for SGNAS on a Tesla V100 GPU. However, FairNAS \cite{fairnas}, GreedyNAS \cite{greedynas}, and ScarletNAS \cite{scarlet} need $48N$, $24N$, and $48N$ GPU hours, respectively, because of the cost of re-executing search. Supernet retraining is needed for FBNetV2 \cite{fbnetv2} and AtomNAS \cite{atomnas}, which makes search very inefficient. 

Note that after finding the best architecture, training from scratch is required in most methods in Table~\ref{table:sota} (including SGNAS, except for AtomNAS \cite{atomnas}). However, training a supernet that can be directly deployed to many constraints (like AtomNAS) needs expensive computation. Even with the time for training from scratch, SGNAS is still more efficient and flexible than AtomNAS. 

\subsection{Experiments on NAS-Bench-201}

\begin{table*}[]
\caption{Performance comparison on different datasets in the NAS-Bench-201 benchmark.}
\centering
\footnotesize
\begin{tabular}{c|c|cc|cc|cc}
\multirow{2}{*}{Method} & \multirow{2}{*}{\begin{tabular}[c]{@{}c@{}}Search Time\\ (GPU hrs)\end{tabular}} & \multicolumn{2}{c|}{CIFAR-10} & \multicolumn{2}{c|}{CIFAR-100} & \multicolumn{2}{c}{ImageNet-16-120} \\ \cline{3-8} 
 &  & Val & Test & Val & Test & Val & Test \\ \hline
optimal & N/A & 91.61 & 94.37 & 73.49 & 73.51 & 46.77 & 47.31 \\ \hline
RSPS \cite{randomsearch} & 2.6 & 84.16$\pm$1.69 & 87.66$\pm$1.69 & 59.00$\pm$4.60 & 58.33$\pm$4.34 & 31.56$\pm$3.28 & 31.14$\pm$3.88 \\
DARTS \cite{darts} & $2.2N$ & 39.77$\pm$0.00 & 54.30$\pm$0.00 & 15.03$\pm$0.00 & 15.61$\pm$0.00 & 16.43$\pm$0.00 & 16.32$\pm$0.00 \\
SETN \cite{setn} & $7.9N$ & 82.25$\pm$5.17 & 86.19$\pm$4.63 & 56.86$\pm$7.59 & 56.87$\pm$7.77 & 32.54$\pm$3.63 & 31.90$\pm$4.07 \\
GDAS \cite{gdas} & $6.6N$ & 90.00$\pm$0.21 & 93.51$\pm$0.13 & \textbf{71.14$\pm$0.27} & \textbf{70.61$\pm$0.26} & 41.70$\pm$1.26 & 41.84$\pm$0.90 \\
\textbf{SGNAS (Ours)} & \textbf{2.5} & \textbf{90.18$\pm$0.31} & \textbf{93.53$\pm$0.12} & 70.28$\pm$1.2 & 70.31$\pm$1.09 & \textbf{44.65$\pm$2.32} & \textbf{44.98$\pm$2.10}
\end{tabular}
\label{table:nas-bench-201}
\end{table*}

To demonstrate efficiency and robustness of SGNAS more fairly, we evaluate it based on a NAS benchmark dataset called NAS-Bench-201 \cite{NAS-Bench-201:}. NAS-Bench-201 includes 15,625 architectures in total. It provides full information of the 15,625 architectures (e.g., top-1 accuracy and FLOPs) on CIFAR-10, CIFAR-100, and ImageNet-16-120 datasets \cite{iamgenet16}, respectively.

Based on the search space defined by NAS-Bench-201, we follow SETN \cite{setn} to train the supernet by uniform sampling. After that, the architecture generator is applied to search architectures on the supernet. We search based on the CIFAR-10 dataset and look up the ground-truth performance of the searched architectures on CIFAR-10, CIFAR-100, and ImageNet-16-120 datasets, respectively. This process is run for three times, and the average performance is calculated as in Table~\ref{table:nas-bench-201}. We see that the architectures searched by SGNAS outperform previous methods on both CIFAR-10 and ImageNet16-120. It is worth noting that, with the supernet training strategy same as SETN \cite{setn}, our result greatly surpasses SETN \cite{setn} on all three datasets. Moreover, the required search time of SGNAS is only 2.5 GPU hours even for $N$ different hardware constraints. 

\begin{figure}[t]
    \centering
    \subfigure[]{
        \begin{minipage}[t]{0.32\linewidth}
        \centering
         \includegraphics[width=\textwidth]{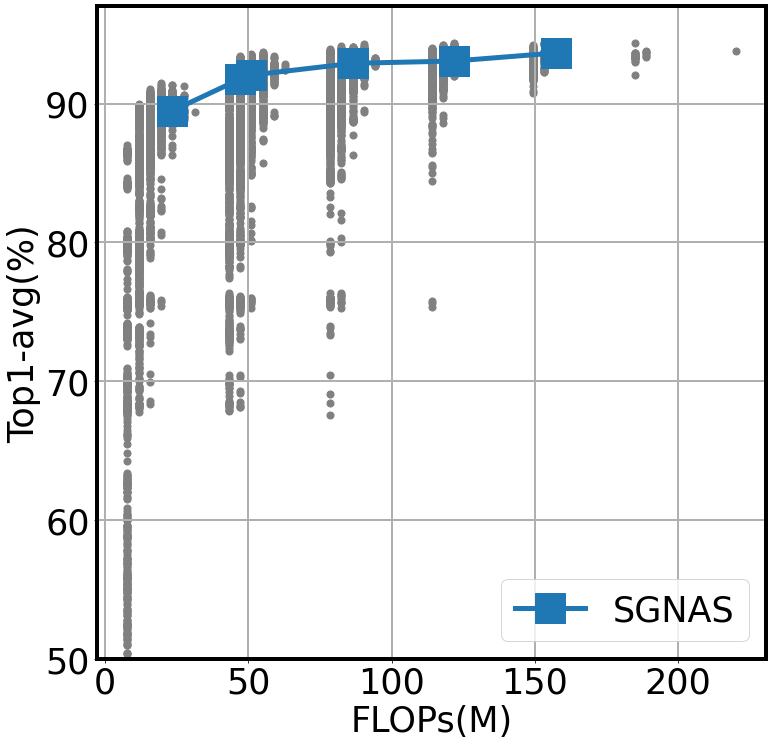}
        \end{minipage}%
    \label{fig:SGNAS_bench_201_cifar10}
    }%
    \subfigure[]{
        \begin{minipage}[t]{0.32\linewidth}
        \centering
         \includegraphics[width=\textwidth]{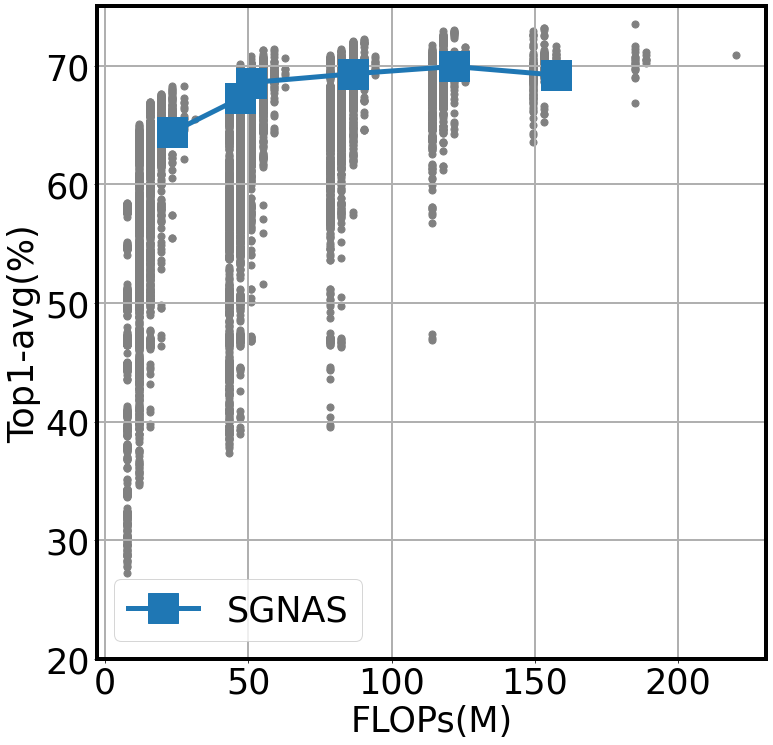}
        \end{minipage}%
    \label{fig:SGNAS_bench_201_cifar100}
    }%
    \subfigure[]{
        \begin{minipage}[t]{0.32\linewidth}
        \centering
         \includegraphics[width=\textwidth]{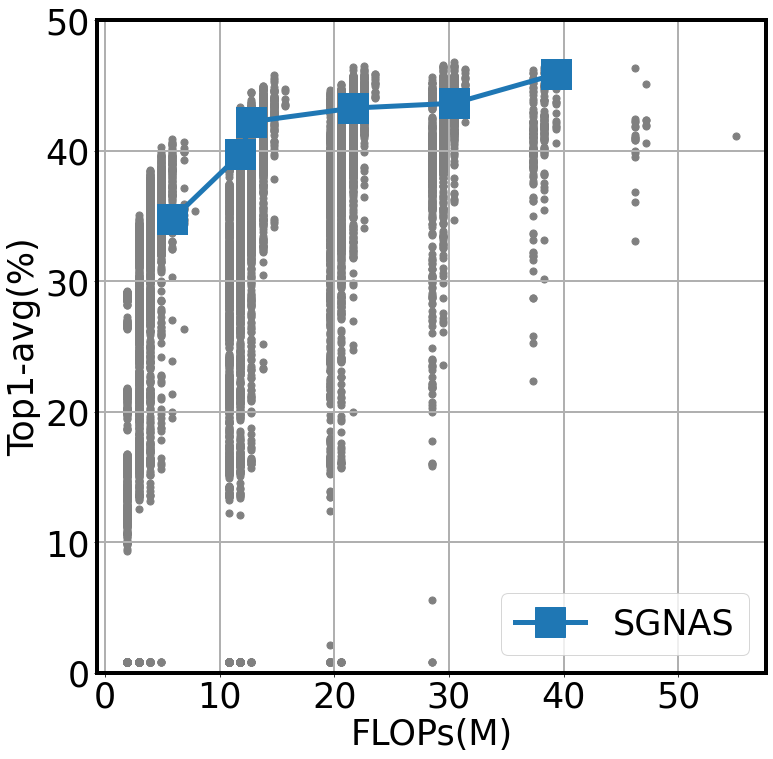}
        \end{minipage}%
    \label{fig:SGNAS_bench_201_imagenet16}
    }%
    \caption{Search results of SGNAS on the CIFAR10, CIFAR100, and ImageNet16-120 datasets. (a) Result on CIFAR-10; (b) Result on CIFAR100; (c) Result on ImageNet16-120.}
 	\label{fig:nas_bench_201}
    \centering
\end{figure}

We show the 15,625 architectures in NAS-Bench-201 on each dataset as gray dots in Fig.~\ref{fig:nas_bench_201}, and draw the architectures searched by the architecture generator under different FLOPs as blue rectangles. After searching once, the architecture generator can generate all blue rectangles directly without re-searching. Moreover, various generated architectures approach the best among all choices. 

\subsection{Performance on Object Detection}
\begin{table}[]
\caption{Performance comparison on the COCO object detection. $^\dagger$: Our implementation result. $^*$ reported in \cite{scarlet}\cite{fairnas}.}
\centering
\footnotesize
\begin{tabular}{c|cc|c}
Model & FLOPs(M) & Top-1 (\%) & mAP (\%) \\ \hline
MobileNetV2$^*$ \cite{mobilenetv2} & 300 & 72.0 & 28.3 \\
MixNet-M$^*$ \cite{mixconv} & 360 & 77.0 & 31.3 \\
FairNAS-A$^*$ \cite{fairnas} & 392 & 77.5 & 32.4 \\
Scarlet-A$^*$ \cite{scarlet} & 365 & 76.9 & 31.4 \\ \hline
MobileNetV2$^\dagger$ & 300 & 72.0 & 29.4 \\
\textbf{SGNAS-A (Ours)} & 373 & 77.1 & 33.9
\end{tabular}
\label{tb:objectdetection}
\end{table}
To verify the transferability of SGNAS on object detection, we adopt the RetinaNet \cite{retinanet} implemented in MMDetection \cite{mmdetection} to do object detection, but replace its backbone by the network searched by SGNAS. The models are trained and evaluated on the MS COCO dataset \cite{coco} (train2017 and val2017, respectively) for 12 epochs with batch size 16 \cite{fairnas}\cite{scarlet}. We use the SGD optimizer with 0.9 momentum and 0.0001 weight decay. The initial learning rate is 0.01, and is multiplied by 0.1 at epochs 8 and 11. Table~\ref{tb:objectdetection} shows that SGNAS has better transferability than the baselines, especially in terms of mAP.

\section{Ablation Studies}
\subsection{Analysis of SGNAS}
\begin{figure}[t]
    \centering
    \subfigure[]{
        \begin{minipage}[t]{0.52\linewidth}
        \centering
         \includegraphics[width=\textwidth]{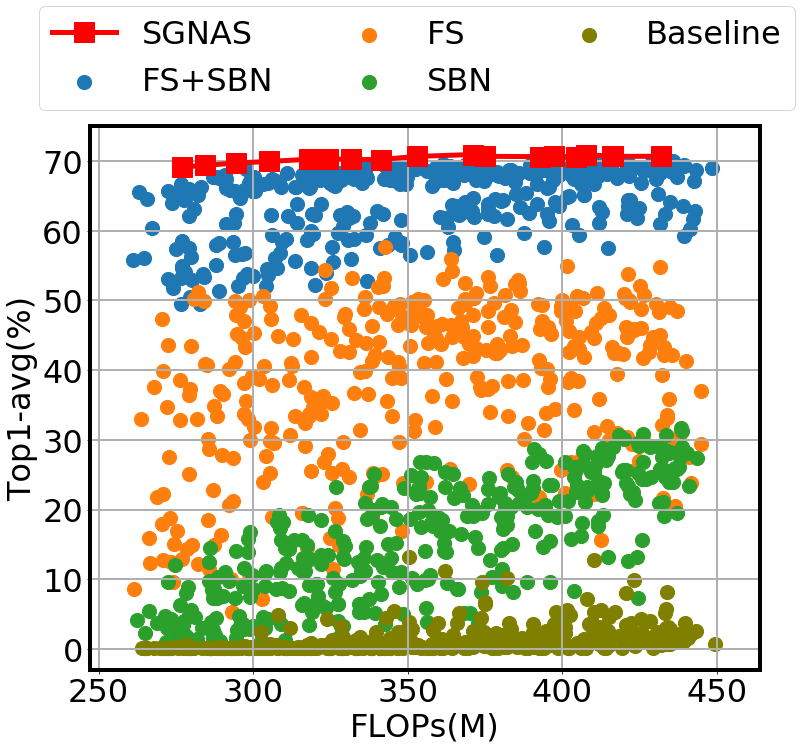}
        \end{minipage}%
    }%
    \subfigure[]{
        \begin{minipage}[t]{0.5\linewidth}
        \centering
         \includegraphics[width=\textwidth]{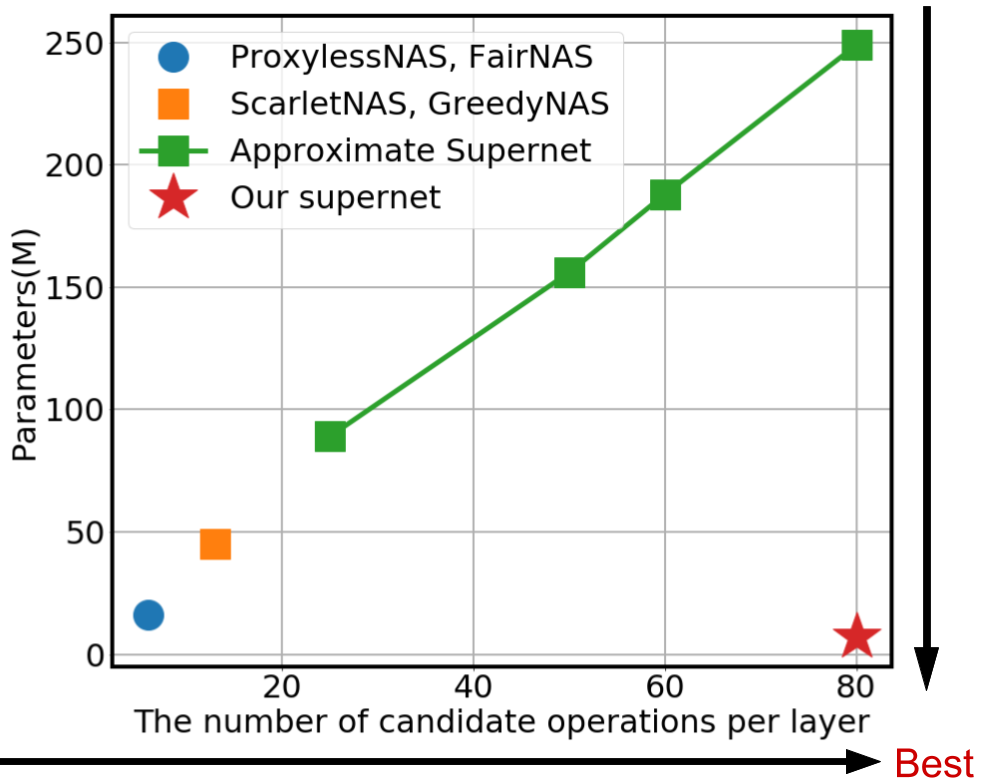}
        \end{minipage}%
    }%
    \caption{(a) Top-1 validation accuracy of randomly sampled architectures (blue dots) and the architectures searched by SGNAS (red rectangles). Performance of other variants is also shown. (b) The relationship between the number of supernet parameters and the number of candidate operations in each layer.}
 	\label{fig:supernet_evaluate}
    \centering
\end{figure}

In Fig.~\ref{fig:supernet_evaluate}(a), we randomly sample 360 architectures from the search space and illustrate the corresponding top-1 validation accuracies as blue dots. Moreover, we draw the architectures searched by SGNAS under different hardware constraint as red rectangles. As can be seen, the architectures searched by SGNAS are almost always the best. 

\subsection{Analysis of Unified Supernet}

\subsubsection{Efficiency of Unified Supernet}

\begin{table}[]
\caption{Comparison in terms of of GPU memory consumption and search time of the architecture generator.}
\vspace{1mm}
\centering
\footnotesize
\begin{tabular}{c|c|c|c|c}
\begin{tabular}[c]{@{}c@{}}Unified \\ Supernet\end{tabular} & \begin{tabular}[c]{@{}c@{}}Search\\ Space\end{tabular} & \begin{tabular}[c]{@{}c@{}}Batch\\ Size\end{tabular} & \begin{tabular}[c]{@{}c@{}}Search Time\\ (GPU hrs)\end{tabular} & \begin{tabular}[c]{@{}c@{}}Memory Cost\\ (GPU)\end{tabular} \\ \hline
\multirow{2}{*}{$\times$} & \multirow{2}{*}{$6^{19}$} & 32 & \multirow{2}{*}{11} & 10.5GB \\
 &  & 128 &  & 40 GB \\ \hline
\multirow{2}{*}{$\checkmark$} & \multirow{2}{*}{\textbf{$80^{19}$}} & 32 & \multirow{2}{*}{\textbf{5}} & \textbf{9GB} \\
 &  & 128 &  & \textbf{28GB}
\end{tabular}
\label{table:super_efficient}
\end{table}

To show efficiency of the proposed unified supernet, we report the relationship between the total number of parameters in the (unified) supernet and the number of candidate operations per layer in Fig.~\ref{fig:supernet_evaluate}(b). For fair comparison, we calculate the number of parameters of different supernets all based on 19 layers. As can be seen, the number of possible operations of the unified supernet in each layer is 7 times larger than GreedyNAS \cite{greedynas} and ScarletNAS \cite{scarlet}, but the number of parameters needed to represent this unified supernet is only 1/6 times of them. The number of possible operations is 13 times larger than FairNAS \cite{fairnas} and ProxylessNAS \cite{proxylessnas}, but the number of supernet parameters for the unified supernet is only half of them. To compare under the same size of search space, we estimate the number of required supernet parameters in previous single-path methods \cite{fairnas}\cite{greedynas}\cite{scarlet}\cite{proxylessnas} when the number of possible operations in each layer increases to 25, 50, 60, and 80, and show them by green squares. Fig.~\ref{fig:supernet_evaluate}(b) shows that the required parameters are significantly boosted when the number of possible operations increases, while the unified supernet avoids this intractability. Under the same size of search space, the number of needed parameters to represent the unified supernet is only 1/35 times of estimated supernets. 

Table~\ref{table:super_efficient} shows the comparison in terms of GPU memory consumption and search time of the architecture generator when it works based on the unified supernet or based on the previous single-path supernet \cite{fairnas}\cite{proxylessnas}. Based on the unified supernet, the GPU memory consumption reduces to 12 GB and the search time is only $0.45$ times of that based on the previous supernet.

\subsubsection{Training Stabilization}
Although the unified supernet largely reduces supernet parameters, the large search space makes the supernet hard to train. To study the effect of force sampling (FS) and shadow batch normalization (SBN) \cite{mixpath} on supernet training, we train the supernet based on different settings, including baseline, with FS, with SBN, and with both FS and SBN. After training the supernet, we randomly sample 360 architectures from the search space and show the corresponding top-1 accuracies in Fig.~\ref{fig:supernet_evaluate}(a). Without FS and SBN, because of large variability and complex architecture, the baseline supernet is hard to train. After utilizing SBN, variability can be well characterized, and the performance becomes more stable. After applying FS, complexity of the supernet is greatly reduced by reducing architecture redundancy. Performance is largely boosted when redundancy is reduced. With both FS and SBN, the unified supernet can more consistently represent architectures with better performance. 

\subsection{Study of Random Priors}

\begin{figure}[t]
    \centering
    \subfigure[]{
        \begin{minipage}[t]{0.49\linewidth}
        \centering
         \includegraphics[width=\textwidth]{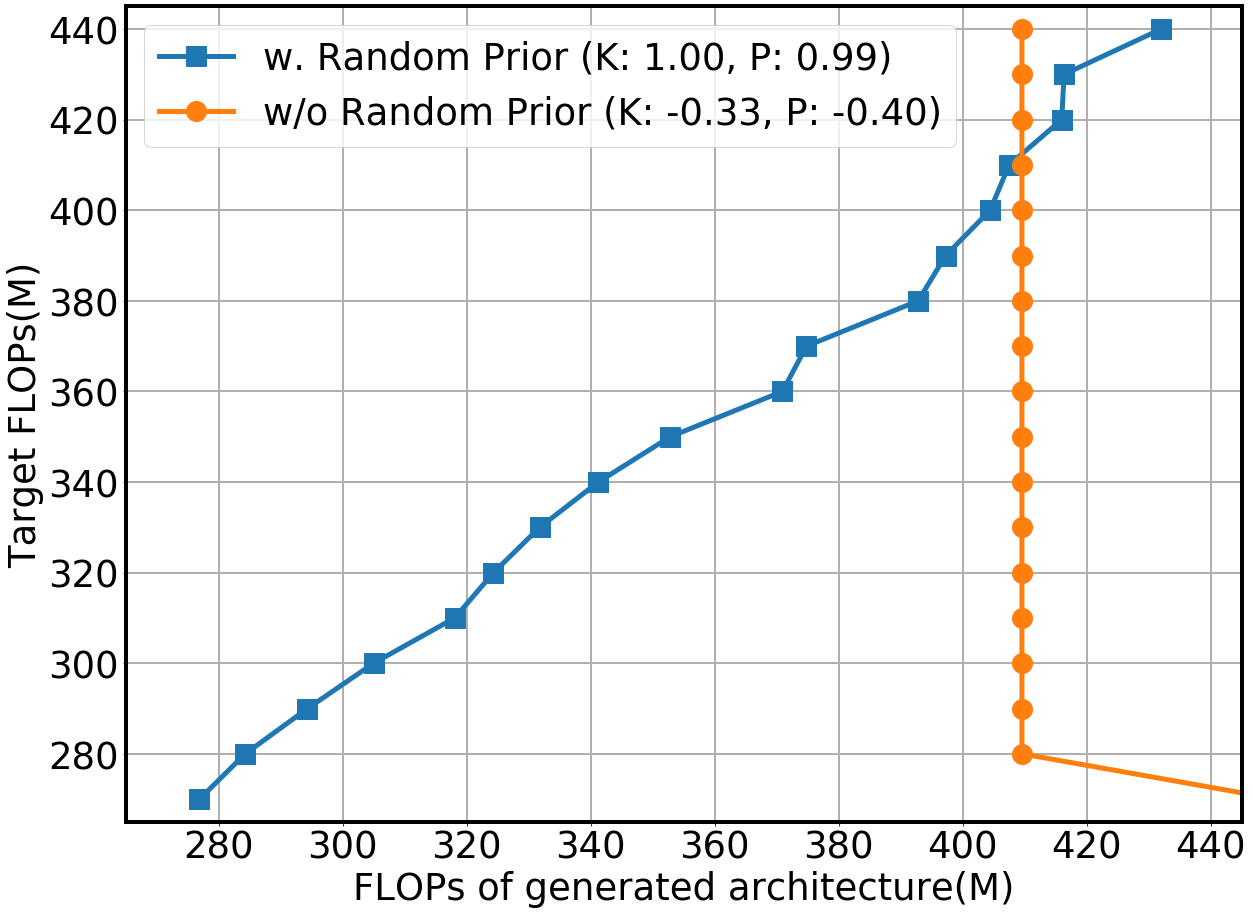}
        \end{minipage}%
    \label{fig:SGNAS_generator_random prior_evaluation}
    }%
    \subfigure[]{
        \begin{minipage}[t]{0.49\linewidth}
        \centering
         \includegraphics[width=\textwidth]{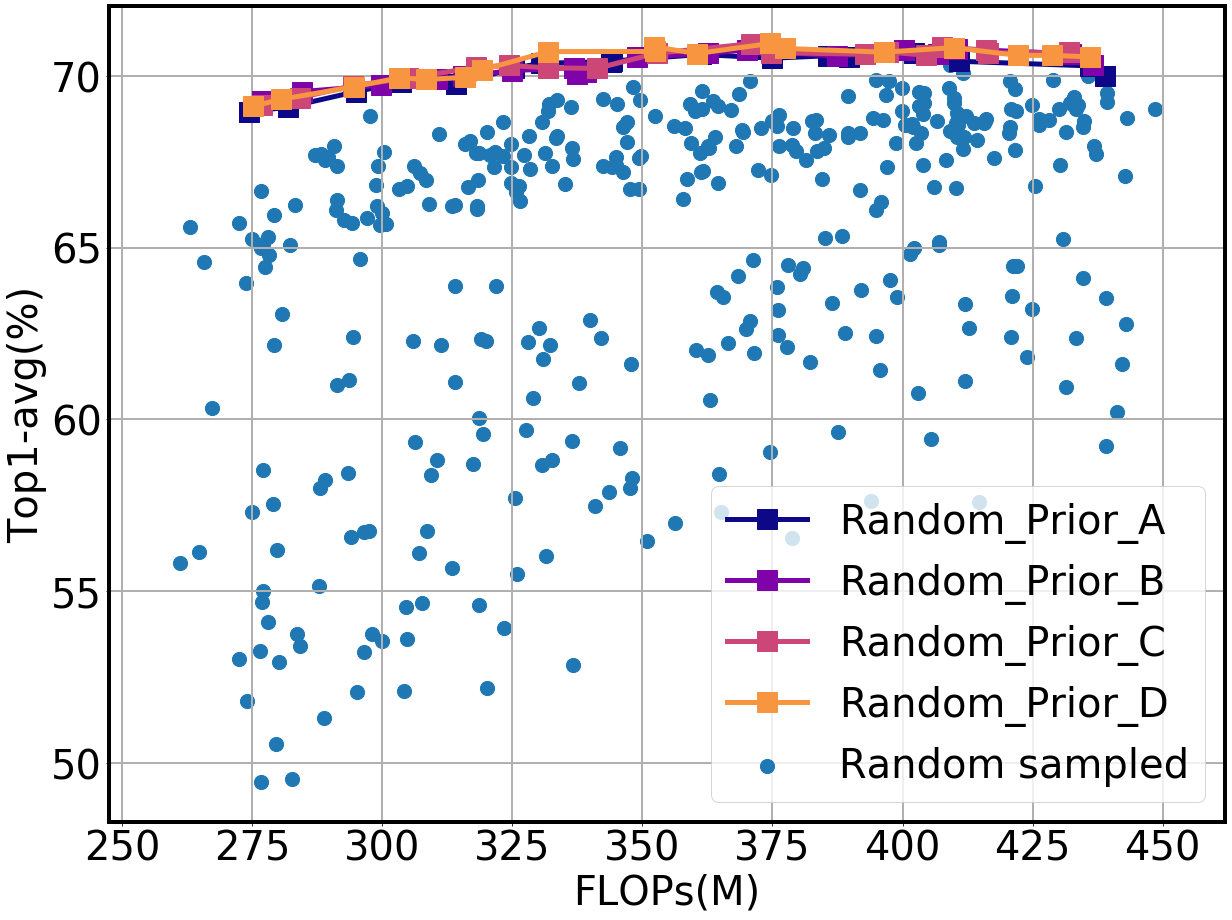}
        \end{minipage}%
    \label{fig:SGNAS_random prior_quality}
    }%
    \caption{(a) The relationship between target FLOPs and the FLOPs of generated architecture. (b) Performance of the architectures randomly sampled from the unified supernet (blue dots) and those generated by the architecture generator trained based on four random priors. }
 	\label{fig:generator_random prior}
    \centering
\end{figure}

To enable the generator to generate architectures under various hardware constraints accurately, random prior is given as the input of the generator. In Fig.~\ref{fig:generator_random prior}(a), we show the correlation between the target FLOPs and the FLOPs of the generated architectures. With the random prior, the generator can generate architectures much more accurately. 
With the random prior, the Kendall tau correlation between the target FLOPs and the generated is 1, while the Pearson correlation is 0.99, which are significantly positive. 

We randomly sample four sub-networks A, B, C, and D from the unified supernet as four priors to train the architecture generator. Inherited from the weights of the unified supernet, the top-1 validation accuracy of these four sub-networks are 58.02\%, 63.36\%, 66.48\%, and 68.48\%, respectively. Fig.~\ref{fig:generator_random prior}(b) shows that, no matter starting from good priors or bad priors, the corresponding trained architecture generators are able to generate the architecture yielding the best performance. This shows that random priors are not to improve the top-1 accuracy, but to give reasonable priors to make the architecture generator generate good architectures under the target constraints. 

\section{Conclusion}
To improve efficiency and flexibility of finding best sub-networks from the supernet under various hardware constraints, we propose the idea of architecture generator that searches the best architecture by generating it. This approach is very efficient and flexible for that only one forward pass is needed to generate good architectures for various constraints, comparing to previous one-shot NAS methods. To ease GPU memory consumption and boost searching, we propose the idea of unified supernet which consists of a stack of unified blocks. We show that the proposed one-shot framework, called SGNAS (searching by generating NAS), is extremely efficient and flexible by comparing with state-of-the-art methods. We further comprehensively investigate the impact of architecture generator and unified supernet from multiple perspectives. Please refer to supplementary materials for the limitation of SGNAS.
    
\textbf{Acknowledgement}
This work was funded in part by Qualcomm through a Taiwan University Research Collaboration Project and in part by the Ministry of Science and Technology, Taiwan, under grants 108-2221-E-006-227-MY3, 107-2923-E-006-009-MY3, and 109-2218-E-002-015. 

{\small
\bibliographystyle{ieee_fullname}
\bibliography{egbib}
}
\newpage
\appendix

\maketitle
\setcounter{figure}{7}
\setcounter{table}{5}    

\section{More Details of Experimental Settings}
\subsection{Dataset}
We perform all experiments based on the ImageNet dataset \cite{imagenet}. Same as the settings in \cite{proxylessnas}\cite{greedynas}\cite{scarlet}, we randomly sample 50,000 images (50 images for each class) from the training set as our validation set, and the rest is kept as the training set. The original validation set is taken as the test set to measure the final performance of each model. The resolution of input images is 224$\times$224. 

\subsection{Supernet Training}
We train the unified supernet for 50 epochs using batch size 256 and adopt the stochastic gradient descent optimizer with a momentum of 0.9 and weight decay of $4 \times 10^{-5}$. The learning rate is decayed based on the cosine annealing strategy from initial value 0.045. We train the unified supernet with strict fairness \cite{fairnas} so that each operation in all sub-blocks and each expansion rate are trained fairly. 

\subsection{Generator Training}
After supernet training, the architecture generator is trained for 50 epochs using batch size 128 by the Adam optimizer with the learning rate 0.001, momentum (0.5, 0.999), and weight decay 0. The temperature $\tau$ of Gumbel Softmax \cite{gumbel_softmax} in Eq.~(2) is
initially set to 5.0 and annealed by a factor of 0.95 for each epoch. The trade-off parameter $\lambda$ in Eq.~(11) is set to 0.0003 in our experiment.

\section{Details of Search Space}
The marco-architecture of our unified supernet is shown in Table~\ref{tb:search_space}. 

\section{More Details of Architecture Generator}
A random prior is encoded into a one-hot format and then is reshaped into the shape of architecture parameters to be generated. The output of the architecture generator is a parameter map with size $Layer Size$ $\times$ $Operation Number$. Motivated by generative adversarial networks, we reshape a random prior so that its shape is the same as the output map. We then feed it to the architecture generator, where we can apply 2D convolution with stride 1 for processing. Without carefully tuning convolution parameters, we can ensure that shape of the output map fits different search spaces. This design is to make the generator easily adapt to different settings. We have experimented with various structures for the architecture generator (e.g., fully connected) and found that convolutional layers yield reliable results.

\begin{table}[]
\caption{Macro-architecture of the search space. MBConv $3\times 3$ denotes MobileNetV2 \cite{mobilenetv2} block with kernel size 3. Column-C denotes the number of output channel of a block. Column-N denotes the number of the blocks. Column-S denotes the stride of the first block when stacked for multiple blocks. Column-E denotes the expansion rate of the blocks, and the tuples of three values represent the lowest value, highest value, and steps between options (low, high, steps).}
\begin{center}
\begin{tabular}{c|c|c|c|c|c}
\hline
Input shape       & Block                 & C & N & S & E \\ \hline
$224^2 \times 3$  & Conv 3$\times$3     & 32  & 1 & 2 & -     \\
$112^2 \times 32$ & MBConv 3$\times$3 & 16  & 1 & 1 & 1     \\ \hline
$112^2 \times 16$ & Unified Block       & 32  & 2 & 2 & (2, 6, 1)      \\
$56^2 \times 32$  & Unified Block       & 40  & 4 & 2 & (2, 6, 1)      \\
$28^2 \times 40$  & Unified Block       & 80  & 4 & 2 & (2, 6, 1)      \\
$14^2 \times 80$  & Unified Block       & 96  & 4 & 1 & (2, 6, 1)      \\
$14^2 \times 96$  & Unified Block       & 192 & 4 & 2 & (2, 6, 1)      \\
$7^2 \times 192$  & Unified Block       & 320 & 1 & 1 & (2, 6, 1)      \\
$7^2 \times 320$  & Unified Block       & 1280& 1 & 1 & (2, 6, 1)      \\ \hline
$7^2 \times 1280$ & Avg pool & -   & 1 & 1  & -    \\
$1280$            & FC       & 1000& 1 & -  & -  \\ \hline
\end{tabular}
\end{center}
\label{tb:search_space}
\end{table}

\begin{figure}
\begin{center}
   \includegraphics[width=1\linewidth]{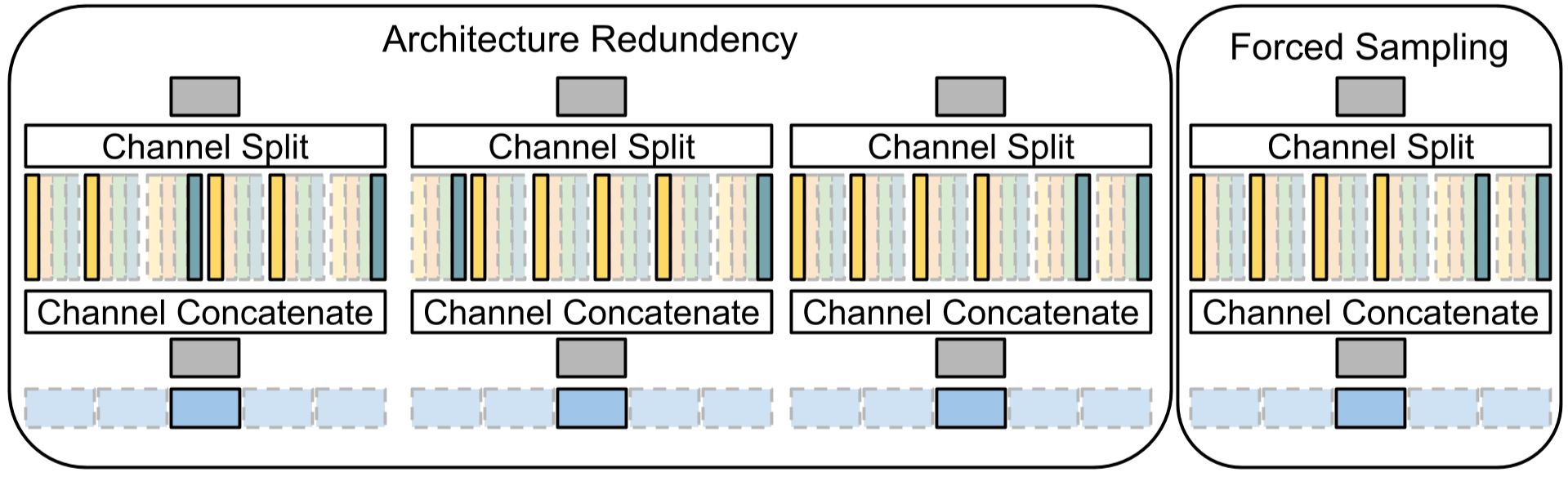}
\end{center}
   \caption{Illustration of architecture redundancy and forced sampling (FS).}
\label{fig:architecture_redundancy}
\end{figure}

\section{More Details of Architecture Redundancy}
We illustrate architecture redundancy in the left of Fig.~\ref{fig:architecture_redundancy} and forced sampling (FS) in the right of Fig.~\ref{fig:architecture_redundancy}. In the four unified blocks in Fig.~\ref{fig:architecture_redundancy}, four depthwise convolution with kernel size $3\times 3$ and two skip connections are used in different sub-blocks. However, the three situations on the left of Fig.~\ref{fig:architecture_redundancy} are treated as different because of different arrangements. With FS, we enforce arrangement of the operations to be unique, and thus only the unified block on the right of Fig.~\ref{fig:architecture_redundancy} can be sampled.

\section{Visualization of Searched Architectures}
\begin{figure*}[t]
\begin{center}
   \includegraphics[width=1\linewidth]{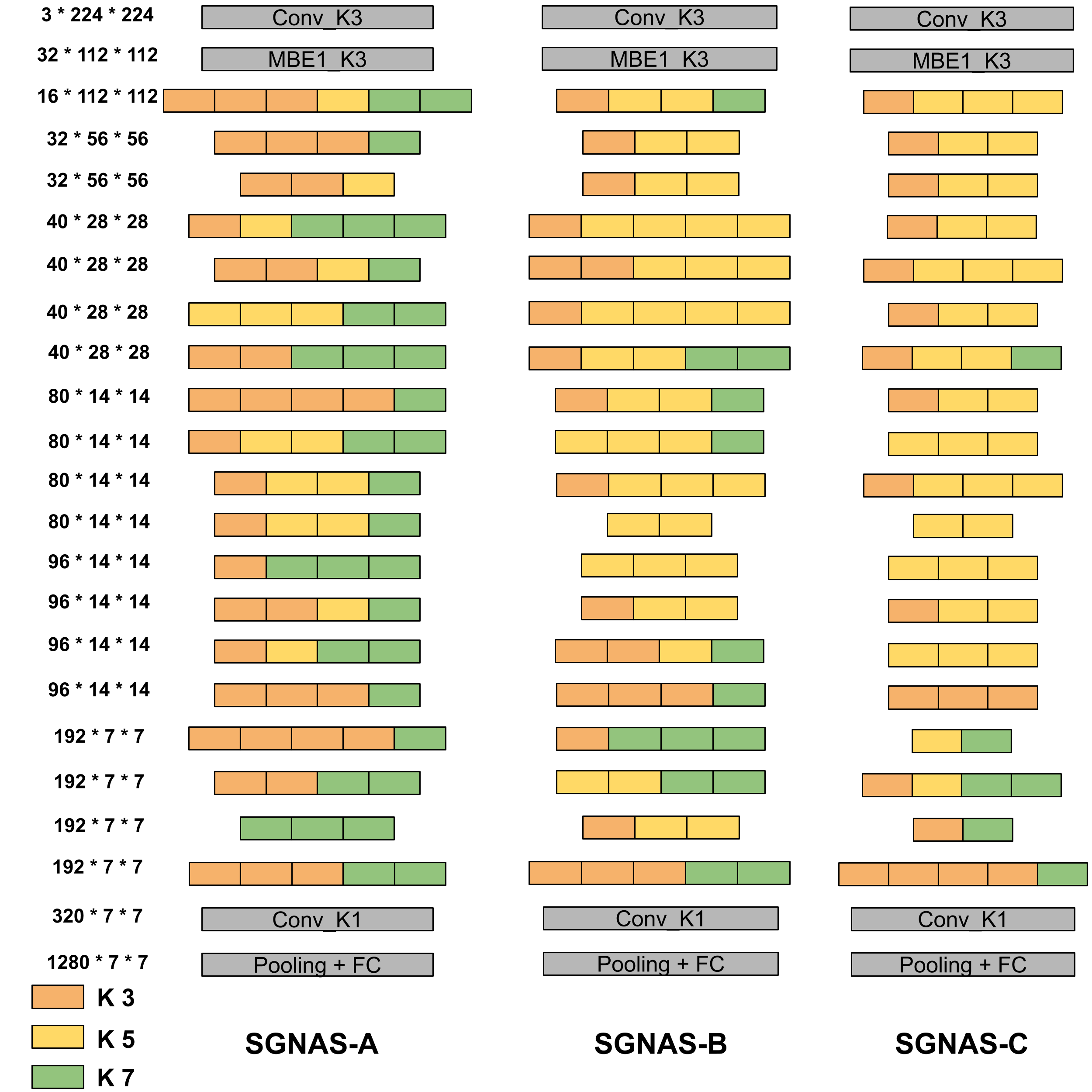}
\end{center}
  \caption{Visualization of the architectures searched by SGNAS (SGNAS-A, SGNAS-B, and SGNAS-C). "MBE1" denotes the mobile inverted bottleneck convolution layers with expansion rate 1. "KX" denotes depthwise convolution with the kernel size X. The gray blocks are predefined blocks before searching.}
\label{fig:architectures}
\end{figure*}

\begin{figure*}[t]
\begin{center}
   \includegraphics[width=1\linewidth]{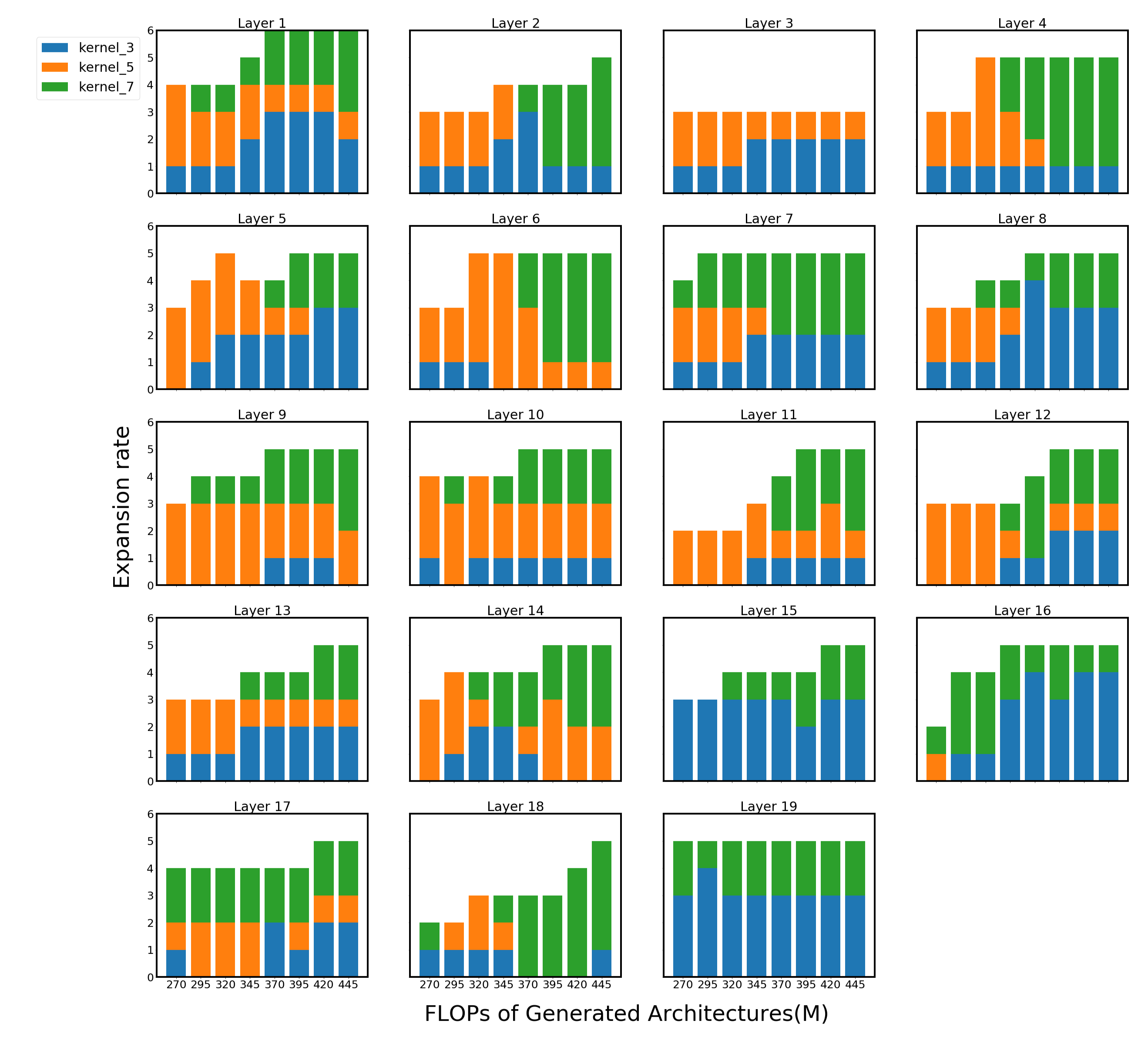}
\end{center}
  \caption{Visualization of the architectures search by SGNAS under different hardware constraints.}
\label{fig:architectures_flow}
\end{figure*}

We visualize SGNAS-A, SGNAS-B, and SGNAS-C in Fig.~\ref{fig:architectures}. Besides, we also visualize the architectures searched by SGNAS under different hardware constraints in Fig.~\ref{fig:architectures_flow}. It is interesting that even if the target hardware constraint is low (e.g., 280M), the expansion rate simulated by sub-blocks is still high in some layers (e.g., layer 1, layer 7, and layer 19).

\section{Limitation}
\textbf{Careful hyperparameter tuning: } In SGNAS, the overall loss function of the architecture generator is defined in Eq.~(10). However, in our experiment, carefully tuning the hyperparameter $\lambda$ for different datasets is required to get trade off between hardware constraints and performance. 

\textbf{Architecture of the architecture generator: } In SGNAS, we manually design architecture of the architecture generator. But we definitely believe that there is a better architecture for the generator. It is worth further study in the future.


\end{document}